\documentclass[10pt,letterpaper]{article}

\usepackage{cogsci}

\cogscifinalcopy 
\usepackage{setspace}
\usepackage{microtype}
\usepackage{graphicx}
\usepackage{subfigure}
\usepackage{booktabs}       
\usepackage{nicefrac}       
\usepackage{float}
\usepackage{hyperref}
\usepackage{algorithm}
\usepackage{amsmath}
\usepackage[table,xcdraw,dvipsnames]{xcolor}
\usepackage{stmaryrd}
\usepackage{amssymb}
\usepackage{todonotes}
\usepackage{comment}
\usepackage{pslatex}
\usepackage{apacite}
\usepackage{acronym}
\usepackage{xspace}
\usepackage{cleveref}
\hypersetup{colorlinks,allcolors=purple}

\makeatletter
\DeclareRobustCommand\onedot{\futurelet\@let@token\@onedot}
\def\@onedot{\ifx\@let@token.\else.\null\fi\xspace}
\def\eg{e.g\onedot} 

\def\ie{i.e\onedot}

\makeatother

\usepackage{CJKutf8}
\newcommand{\zhsc}[1]{\begin{CJK*}{UTF8}{gbsn}#1\end{CJK*}}
\newcommand{\zhtc}[1]{\begin{CJK*}{UTF8}{bsmi}#1\end{CJK*}}

\newcounter{program}
\newenvironment{program}
  {\refstepcounter{program}\equation\small\aligned}
  {\endaligned\endequation}
  
\crefname{program}{program}{programs}
\Crefname{program}{Program}{Programs}

\author{{\large \textbf{Guangyuan Jiang}\normalfont{\textsuperscript{1,3,$\star$}},
    \textbf{Matthias Hofer}\textsuperscript{1},
    \textbf{Jiayuan Mao}\textsuperscript{2},
    \textbf{Lionel Wong}\textsuperscript{1},}\\
    {\large \textbf{Joshua B. Tenenbaum}\textsuperscript{1,2}, \and
    \textbf{Roger P. Levy}\textsuperscript{1}}\\ 
    \textsuperscript{1}MIT BCS \quad \textsuperscript{2}MIT CSAIL \quad \textsuperscript{3}Peking University\\
    $^\star$correspondence to \texttt{jianggy@mit.edu}
}

\acrodef{mdl}[MDL]{minimum description length}
\acrodef{dsl}[DSL]{domain-specific language}

\definecolor{blue1}{RGB}{63,169,245}
\definecolor{blue2}{RGB}{30,76,124}
\definecolor{green1}{RGB}{122,201,67}
\definecolor{orange1}{RGB}{255,147,30}
\definecolor{pink1}{RGB}{253,112,161}
\definecolor{pcc}{RGB}{253,112,161}
\definecolor{mdlc}{RGB}{58,109,145}
\definecolor{dlc}{RGB}{87,159,112}
\definecolor{lsc}{RGB}{208,192,88}
\definecolor{crc}{RGB}{140,182,96}

\newcommand{\myparagraph}[1]{\paragraph{#1}}

\title{Finding structure in logographic writing with library learning}
\begin{document}

\maketitle

\begin{abstract}
One hallmark of human language is its combinatoriality---reusing a relatively small inventory of building blocks to create a far larger inventory of increasingly complex structures. 
In this paper, we explore the idea that combinatoriality in language reflects a human inductive bias toward representational efficiency in symbol systems. We develop a computational framework for discovering structure in a writing system. Built on top of state-of-the-art library learning and program synthesis techniques, our computational framework discovers known linguistic structures in the Chinese writing system and reveals how the system evolves towards simplification under pressures for representational efficiency. We demonstrate how a library learning approach, utilizing learned abstractions and compression, may help reveal the fundamental computational principles that underlie the creation of combinatorial structures in human cognition, and offer broader insights into the evolution of efficient communication systems.

\textbf{Keywords:} 
language learning; evolution; phonology; sketch understanding; bayesian modeling
\end{abstract}

\setstretch{1}
\section{Introduction}
\begin{figure}[t!]
    \centering
    \includegraphics[width=\linewidth]{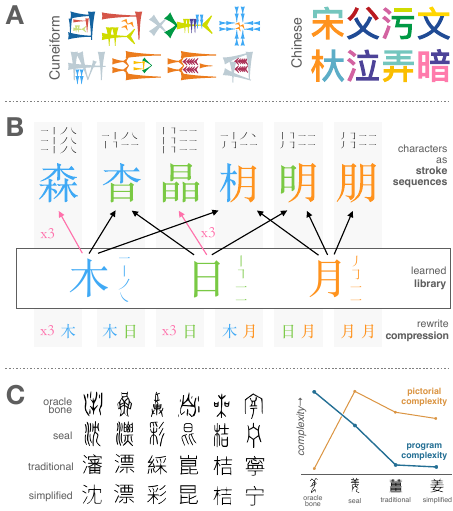}
    \caption{\textbf{An overview of our library learning model for writing systems}:
    \textbf{(A)} Parts are frequently reused (marked in the same color) within and across characters in the multiple logographic writing systems (\eg, Cuneiform, Chinese). 
    \textbf{(B)} In our library learning model, we represent characters as stroke sequences. Learned library functions identify and represent reused parts (\eg, \textcolor{blue1}{\texttt{\zhsc{木}}}) and relations (\eg, \textcolor{pink1}{\texttt{x3}}, repeating three times), leading to program compression and the discovery of structures.
    \textbf{(C)} The model scales to study the multiple scripts in the Chinese writing system across time, revealing trends and adaptations in the use of radicals and other elements.}
    \label{fig:reuse-example}
\end{figure}

Human language is fundamentally combinatorial, reusing a relatively small inventory of building blocks to create a far larger inventory of increasingly complex structures.
This structure is deeply rooted and evident in language's earliest written records, such as cuneiform and oracle bone scripts (\Cref{fig:reuse-example}A,C).
It manifests at multiple levels of structure, from phonemes to words and sentences, and across various linguistic modalities---spoken, signed, and written---illustrating a universal trait of human communication systems \cite{hockett1960origin,zuidema2018evolution}.

In this paper, we explore the idea that combinatoriality in language reflects a human inductive bias toward representational efficiency in symbol systems. Research has recognized that the evolution and structure of these systems are profoundly influenced by an inductive bias towards representational efficiency \cite{kirby2015compression,gibson2019efficiency}, which combinatoriality offers \cite{kirby2022cumulative}.
Empirical support for this notion has come from laboratory experiments \cite{verhoef2014emergence,little2017signal,hofer2019iconicity}, as well as qualitative analyses of emerging sign languages \cite{sandler2011gradual} and logographic writing systems, posited to have evolved towards increasing levels of simplicity \cite{sampson1985writing}. Representational efficiency-based methods have been used for morphology \cite{goldsmith2001unsupervised} and syntax \cite{carroll1992two,kim2019compound}, but structure discovery for these levels of linguistic representation has proven technically challenging.

Here, we develop an efficiency-based structure discovery method for a complex logographic writing system, namely Chinese orthography, using the framework of \textbf{library learning} from the program synthesis literature \cite{ellis2022synthesizing,bowers2023top}. The Chinese orthography presents a unique opportunity for studying combinatorial structure due to their long evolutionary history \cite<of over 3,000 years, see, e.g.,>[]{kane2006chinese} and the frequent reuse of graphical elements (hundreds of radicals) within individual characters and across the writing system. The library learning framework iteratively identifies recurring patterns and stores them in a library of abstractions (\Cref{fig:reuse-example}B). The characters themselves are then redescribed by reference to these abstractions, increasing the overall concision of representation of the character inventory. Applying library learning to the Chinese writing system offers two related opportunities: to validate whether library learning recovers its radical combinatorial structure, and to use library learning to analyze the changes in the system over time.

Reflecting these two opportunities, our investigation consists of two parts.
Part I develops a library learning model as a candidate hypothesis about human inductive biases for combinatorial structure. We validate the model by showing that it successfully rediscovers known linguistic structures, such as radicals, and characters' hierarchical decomposition, of the Chinese writing system in its simplified form.

Part II extends this analysis diachronically, investigating the evolution of Chinese scripts over several representative historical stages. Previous work in emerging sign languages has suggested the hypothesis that languages evolve towards simplification under pressures for representational efficiency \cite{motamedi2019evolving,brentari2017language}. Here, the ancient Chinese writing system allows us to test this hypothesis at a larger scale. However, providing quantitative evidence presents significant challenges, especially when it comes to describing the graphical components. A recent study by \citeA{han2022simplification}, for instance, aimed to evaluate the evolution of the Chinese writing system using a simple measure of pictorial complexity that does not account for internal combinatorial structure within characters, and reported findings that ostensibly conflict with the expected trend toward the hypothesis and previous qualitative arguments in the Chinese writing system \cite{woon1987chinese,sampson1985writing,wang1973chinese,zhao2007planning}. Our result challenges the pictorial complexity and supports earlier qualitative results, suggesting a trend towards simplification over time.

Our findings demonstrate how a library learning approach, utilizing learned abstractions and compression, may help reveal the fundamental computational principles that underlie the creation of combinatorial structures in human cognition, and offer broader insights into the evolution of efficient communication systems.

\begin{figure*}[t]
    \centering
    \includegraphics[width=\linewidth]{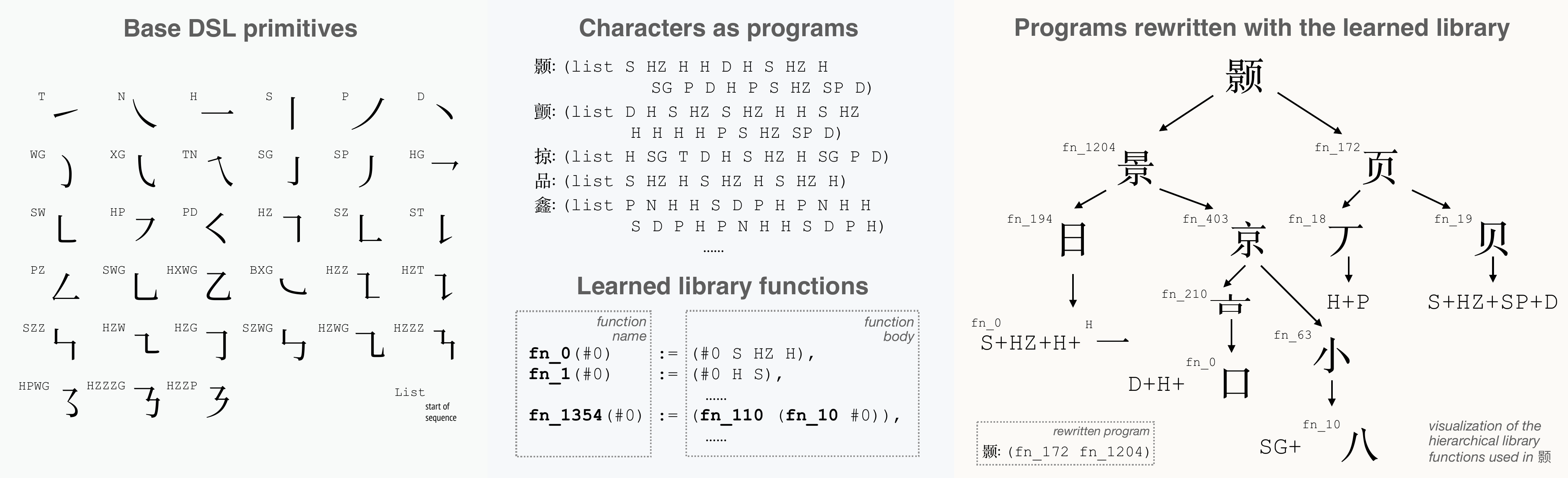}
    \caption{\textbf{(Left)}: Primitives used in the base \ac{dsl} $\mathcal{L}_{base}$, including 33 stroke primitives and one \texttt{list} symbol. \textbf{(Middle)}: Character programs represented in $\mathcal{L}_{base}$ and the library functions learned by the model. \textbf{(Right)}: Visualization of an example character \zhsc{颢}'s hierarchical decomposition discovered, represented as a tree of library functions.}
    \label{fig:banner}
\end{figure*}

\section{Part I: A library learning model for the Chinese writing system}

In this section, we build a library learning-based computational model for writing systems and present studies on the most widely used script of the Chinese writing system: simplified Chinese. Our goal is to leverage library learning to reverse-engineer the core human inductive biases for combinatorial structure in language. We will show that our library learning model can accurately capture the core structural aspects of simplified Chinese, aligned with prior empirical findings and theories about hierarchical decomposition and radicals in the Chinese language.

\subsection{Methods}
We start by describing the computational model designed to learn library functions for the Chinese writing system. The high-level goal of the model is to find efficient representations (good compression in terms of minimal description length, or \acs{mdl}) of all the characters in a writing system. 

\myparagraph{Defining a \texorpdfstring{DSL}{\ac{dsl}}.} 
We start with defining a \acf{dsl} for the Chinese writing system. It enables us to represent individual characters as sequences of primitive strokes. 
This approach is founded on three crucial insights: firstly, Chinese characters consist of strokes; second, these strokes can be classified into a finite number of basic types; and third, the spatial relationships between consecutive strokes are generally consistent. 

Our base \ac{dsl} $\mathcal{L}_{base}$ (visualized in \Cref{fig:banner} left) consists of 33 stroke primitives \{\texttt{T}, \texttt{N}, \texttt{H}, \dots, \texttt{HZZP}\} (six basic strokes + 27 turning strokes, as defined in Chinese language authoritative standard \cite{GF2001-2001}) and one helper primitive \texttt{list} indicating the beginning of the stokes sequence\footnote{Following conventions in $\lambda$-calculus, \texttt{(list A B)} represents a list of two elements, and \texttt{((list A B) C)} is equivalent to a (flattened) list representation \texttt{(list A B C)}. We do not consider nested lists in this paper.}. 
Each stroke primitive in the DSL represents a stroke type by its visual form and how it is written. For modern Chinese characters, we can uniquely encode every character to a LISP-style program $p_\text{char}$ of the stroke sequence based on canonical stroke orderings \cite{GF0023-2020}. For example, for the two-character word \zhsc{认知}, which means ``cognition''. The corresponding program encoding each characters are: $p_\text{\zhsc{认}} := \texttt{(list D HZT SP N)}$ and $p_\text{\zhsc{知}} := \texttt{(list P H H SP N S HZ H)}$.

\myparagraph{Objectives for library learning and the measure of program complexity.}
We formalize the process of learning an efficient representation of a writing system as a library learning problem. At a high level, the library learning algorithm identifies instances of reuse and abstraction from the dataset, facilitating efficient data compression. 
Our algorithm takes as input a set of (literal) character programs ${P}_{\mathcal{L}_{base}}(\mathcal{W}) = \{p_\text{\zhsc{认}}, p_\text{\zhsc{知}}, \dots\}$, with individual programs $p$ represented with primitives in $\mathcal{L}_{base}$. 
The algorithm iteratively infers an optimal library of functions $\mathcal{L}_{*}$ that includes reusable components (abstractions) learned from the programs in addition to the base \ac{dsl}. The programs set ${P}_{\mathcal{L}_{base}}(\mathcal{W})$ rewritten with the new library $\mathcal{L}$ yield ${P}_{\mathcal{L}}(\mathcal{W}) := \{\textsc{Rewrite}(p, \mathcal{L}) \mid p \in P_{\mathcal{L}_{base}}(\mathcal{W})\}$, where $\textsc{Rewrite}(p, \mathcal{L})$, informally, rewrites the program $p$ in the most compact way using functions from the library $\mathcal{L}$.

How do we know which library is optimal? When would the compression algorithm stop? The optimal library $\mathcal{L}_{*}$ for all character programs ${P}_{\mathcal{L}_{base}}$ in a writing system should have a maximum reduction in description length after being rewritten with $\mathcal{L}_*$, while keeping the $\mathcal{L}_{*}$ size overhead small. Following previous program synthesis approaches in bootstrap library learning (\ie, DreamCoder \cite{ellis2022synthesizing} and Stitch \cite{bowers2023top}), we define the program (\acs{mdl}) complexity of writing systems as the description length of the optimal library and the individual programs rewritten with the optimal library. This aligns with the concept of \ac{mdl}: the description length of the shortest program that can generate the targeted data. 

Formally, we define the description length $\mathrm{DL}_{\mathcal{L}}(\mathcal{W})$ of a writing system $\mathcal{W}$ under a specific library $\mathcal{L}$, the description length of the library itself $\mathrm{DL}\left(\mathcal{L}\right)$, and the program (\ac{mdl}) complexity of the writing system, represented as $C(\mathcal{W})$ as follows:
\begin{equation}
    \small
    \begin{aligned}
    \mathrm{DL}_{\mathcal{L}} (\mathcal{W}) &= \overbrace{\sum_{p \in P_{\mathcal{L}_{base}}(\mathcal{W})}\mathrm{DL}\left(\textsc{Rewrite}(p, \mathcal{L})\right)}^{\substack{\text{description length} \\ \text{of the rewritten characters}}}\quad + \overbrace{\vphantom{\sum_{p \in \mathcal{W}}\mathrm{DL}}\mathrm{DL}\left(\mathcal{L}\right)}^{\substack{\text{description length} \\ \text{of the library}}} \\
    \mathrm{DL}\left(\mathcal{L}\right) &= \sum_{\texttt{fn} \in \mathcal{L}}\mathrm{DL}\left(\textsc{Body}(\texttt{fn})\right) \\
    C(\mathcal{W}) &= \min_{\mathcal{L}} \mathrm{DL}_{\mathcal{L}} (\mathcal{W})
    \end{aligned}
    \label{eq:objectives}
\end{equation}

Here, formally, $\textsc{Rewrite}(\cdot, \cdot)$ is an operation that seeks the shortest representation (in terms of $\mathrm{DL}$) to rewrite the literal program using functions from the new library. $\textsc{Body}(\cdot)$ is the program string that defines a library function's body. $\mathrm{DL}(\cdot)$ is a measure of the description length of a program string (which generally reflects the number of functions used); we leverage the exact $\lambda$-calculus-based measure of program size used in Stitch and DreamCoder, specifically the $\mathrm{cost}(\cdot)$ defined in \citeA{bowers2023top}.

\myparagraph{Iterative learning of the library.}
The library learning model's goal is to find the most compressive and concise $\mathcal{L} = \mathcal{L}_{*}$ for the writing system (\ie, minimizing $\mathrm{DL}_{\mathcal{L}} (\mathcal{W})$). 

Specifically, as we are running large-scale library learning on thousands of character programs, we leverage the Stitch algorithm \cite{bowers2023top} for efficiently discovering library functions. Based on state-of-the-art program synthesis techniques, Stitch iteratively performs top-down searches to discover $\lambda$-abstractions \cite{sep-lambda-calculus} as library functions, thereby compressing the programs. Formally, it gradually grows $\mathcal{L}_{base}$ to find $\mathcal{L}_{*}$ by minimizing $\mathrm{DL}_{L}(\mathcal{W})$ (as shown in \cref{eq:objectives}). Compared to the DreamCoder compression algorithm, it is three orders of magnitude faster, making it tractable for us to examine the entire writing system.

Here, a minimal ``writing system'' containing three characters $\mathcal{W} = \{\text{\zhsc{旦}},\text{\zhsc{见}},\text{\zhsc{日}}\}$ will be used as a working example. The initial programs ${P}_{\mathcal{L}_{base}}(\mathcal{W}) = \{p_\text{\zhsc{旦}}, p_\text{\zhsc{见}}, p_\text{\zhsc{日}}\}$ are defined as follows: 

\begin{program}\label{program:danjianri_base}
    p_\text{\zhsc{旦}} &:= \textcolor{CornflowerBlue}{\texttt{(list S HZ}} \texttt{ H H H)}\\
    p_\text{\zhsc{见}} &:= \textcolor{CornflowerBlue}{\texttt{(list S HZ}} \texttt{ SP SWG)}\\
    p_\text{\zhsc{日}} &:= \textcolor{CornflowerBlue}{\texttt{(list S HZ}} \texttt{ H H)}
\end{program}

In the first iteration, the algorithm discovers a (largest) reusable part across all three programs \textcolor{CornflowerBlue}{\texttt{(list S HZ)}}, this part is thus added to the $\mathcal{L}_{base}$ as a library function, yielding a new library $\mathcal{L}_1 := \mathcal{L}_{base} \cup \{\texttt{fn\_0}\}$, where $\textsc{Body}(\texttt{fn\_0}) = \textcolor{CornflowerBlue}{\texttt{(list S HZ)}}$. The character programs in the set ${P}_{\mathcal{L}_{1}}(\mathcal{W})$, rewritten with the new library $\mathcal{L}_1$, are as follows:

\begin{program}\label{program:danjianri_1}
    \textsc{Rewrite}(p_\text{\zhsc{旦}}, \mathcal{L}_1) &= \textcolor{purple}{\texttt{(fn\_0 H H}} \texttt{ H})\\
    \textsc{Rewrite}(p_\text{\zhsc{见}}, \mathcal{L}_1) &= \texttt{(fn\_0 SP SWG)}\\
    \textsc{Rewrite}(p_\text{\zhsc{日}}, \mathcal{L}_1) &= \textcolor{purple}{\texttt{(fn\_0 H H)}}
\end{program}

In the second iteration, the algorithm discovers \textcolor{purple}{\texttt{(fn\_0 H H)}} that is reused and adds it to the library, resulting $\mathcal{L}_2 := \mathcal{L}_{1} \cup \{\texttt{fn\_1}\}$, where $\textsc{Body}(\texttt{fn\_1}) = \textcolor{purple}{\texttt{(fn\_0 H H)}}$.

\begin{program}\label{program:danjianri_2}
    \textsc{Rewrite}(p_\text{\zhsc{旦}}, \mathcal{L}_2) &= \texttt{(fn\_1 H)}\\
    \textsc{Rewrite}(p_\text{\zhsc{见}}, \mathcal{L}_2) &= \texttt{(fn\_0 SP SWG)}\\
    \textsc{Rewrite}(p_\text{\zhsc{日}}, \mathcal{L}_2) &= \texttt{fn\_1}
\end{program}

In this example, $\mathcal{L}_2$ is the optimal library ($\mathcal{L}_{*}$) for this \zhsc{旦见日} writing system. It is also a hierarchical library, as $\texttt{fn\_1}$ is defined based on other library functions $\texttt{fn\_0}$. The program strings in $P_{\mathcal{L}_*}(\mathcal{W})$, rewritten with the optimal library $\mathcal{L}_{*}$, lead to a more concise (compressed) description of the writing system (\cref{program:danjianri_2}).

\begin{figure}[t]
    \centering
    \includegraphics[width=0.95\linewidth]{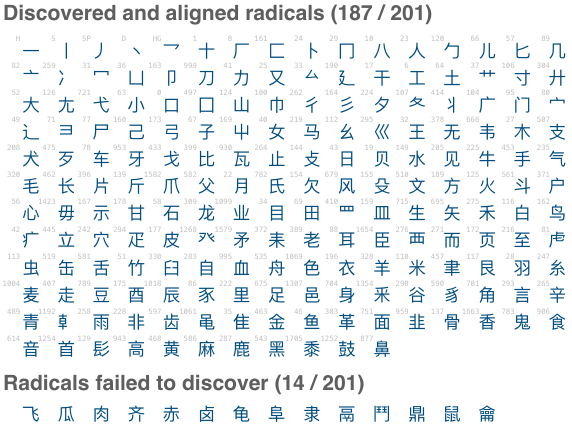}
    \caption{\textbf{Visualization of the aligned MoE radical--library function pairs.} 201 radicals from the MoE radicals set are colored in \textcolor{blue2}{blue}, corresponding library functions are colored in \textcolor{gray}{gray} on the top left (we omit the \texttt{fn\_} prefix for brevity). Our model discovered most of the expert-defined radicals ($93.0\%$).}
    \label{fig:moe_aligned}
\end{figure}

\myparagraph{Data.}
A total of 6,596 simplified Chinese characters were collected, including their canonical stroke decompositions and orderings, from the Han character library \cite{cjklib}. This dataset covers a majority of the official set of simplified Chinese characters (6,596 out of 6,763) specified in the \citeA{GB2312} standard.

\subsection{Results}
\myparagraph{Learned library captures reuse and hierarchical decomposition in the Chinese writing system.}

Inspired by previous work on using library learning to discover geometric shape patterns \cite{sable2022language} and object structures \cite{wong2022identifying}, we applied our model to analyze the library functions it learned from 6,596 simplified Chinese characters. 

Our model discovered a set of hierarchical library functions for compressing the writing system. In addition to the 34 primitives defined in the base \ac{dsl} $\mathcal{L}_{base}$, 1,805 learned library functions were found in the optimal library $\mathcal{L}_{*}$, resulting in a library size of $\lvert \mathcal{L}_{*} \rvert = 1,839$. 
Among all the new library functions learned, 1,717 ($95.1\%$) were hierarchically defined, meaning that they were not only composed of base DSL primitives but also incorporated other newly learned library functions (see \Cref{tab:example_library} for examples). 

This hierarchical organization facilitated significant compression of the writing system, achieving an overall compression rate of $4.16\times$, defined by $\mathrm{DL}_{\mathcal{L}_{base}}(\mathcal{W}) / \mathrm{DL}_{\mathcal{L}_{*}}(\mathcal{W})$.
In terms of individual characters, the program description length was drastically reduced. On average, the number of functions required to encode an individual character was $1/5.63$ of the original amount. As shown in the right of \Cref{fig:banner}, \zhsc{颢} (pronounced as: \textit{hào}, meaning \textit{bright sunlight}) is represented by a program containing only two functions in $\mathcal{L}_{*}$ compared to the 19 functions required in $\mathcal{L}_{base}$ before compression. Delving into the two functions learned to encode \zhsc{颢}, \texttt{fn\_1204} resembles \zhsc{景} and \texttt{fn\_172} \zhsc{页}, which are both commonly used in several other characters (\eg, \zhsc{惊}, \zhsc{影}). The \texttt{fn\_1204} (\zhsc{景}) and \texttt{fn\_172} (\zhsc{页}) were also hierarchically defined on other functions (\ie, \zhsc{日}, \zhsc{京}, \zhsc{贝}). This contributes to an explicit hierarchical representation that captures reuse and structure at a system level for simplified Chinese.

The example of \zhsc{颢} intuitively demonstrates that our model's learned hierarchical decomposition closely resembles the way humans cognitively break down structures. To assess the validity of these decompositions, we compared the model's predictions against a gold standard from the Han character library. The comparison, based on 3,052 characters, revealed that the model's hierarchical representations achieved an overall recall rate of $76.3\%$ and an F$_1$ score of $61.6$ over the spans of the parsed trees (results are shown in \Cref{tab:parsing_evaluation}). 
It is important to emphasize the significance of the recall (true positive) rate in this context, as our model prioritizes maximum compression, potentially uncovering more fine-grained patterns of reuse than the ones typically recognized by humans.
These findings suggest that our model effectively captures the intrinsic hierarchical organization of simplified Chinese characters.

\begin{table}[thbp!]
    \centering
    \resizebox{\linewidth}{!}{%
    \begin{tabular}{l@{\hspace{0.3em}}l|l@{\hspace{0.3em}}lll}
        \toprule
        \multicolumn{2}{l|}{Library function discovered}          & \#Uses  &  (percentile)          & Semantic & Example usage  \\
        \midrule
        \texttt{fn\_0(\#0)  }  &\texttt{:= (\#0 S HZ H)}         & 3342    &  ($100\%$)          & \zhsc{口}        & \zhsc{口品扣亩}   \\
        \texttt{fn\_23(\#0) }  &\texttt{:= (\#0 SP N)}           & 418     &  ($99\%$)          & \zhsc{人}        & \zhsc{人仄灭认}   \\
        \texttt{fn\_48(\#0) }  &\texttt{:= (\#0 H SG)}           & 220     &  ($98\%$)          & \zhsc{夫}        & \zhsc{撵潜窥肤}   \\
        \texttt{fn\_105(\#0)}  &\texttt{:= (fn\_23 (fn\_2 \#0))} & 83      &  ($95\%$)          & \zhsc{寸}        & \zhsc{夺守尊将}   \\
        \texttt{fn\_209(\#0)}  &\texttt{:= (fn\_66 (fn\_0 \#0))} & 38      &  ($90\%$)          & \zhsc{兄}        & \zhsc{兑兕党况}   \\
        \texttt{fn\_415(\#0)}  &\texttt{:= (fn\_20 (fn\_3 \#0))} & 15      &  ($80\%$)          & \zhsc{喜}        & \zhsc{僖嘻嬉喜}   \\
        \texttt{fn\_776()   }  &\texttt{:= (fn\_128 fn\_274)}    & 6       &  ($60\%$)          & \zhsc{比}        & \zhsc{皆比毕毖}   \\
        \texttt{fn\_1624()  }  &\texttt{:= (fn\_1233 T)}         & 2       &  ($20\%$)          & \zhsc{禺}        & \zhsc{愚遇}      \\
        \bottomrule
    \end{tabular}
    }
    \caption{\textbf{Examples of learned library functions}, ordered according to their frequency of usage. \texttt{(\#}$\cdot$\texttt{)} indicates the parameter a function takes in.} \label{tab:example_library}
    \vspace{-0.5em}
\end{table}

\begin{table}[thbp!]
    \centering
    \scriptsize
    \begin{tabular}{l|p{0.6cm}p{0.6cm}p{0.6cm}p{1.2cm}}
        \toprule
        \multicolumn{1}{l|}{Model}        & F$_1$ & Precision & Recall & Exact match \\
        \midrule
        Library learning                 & 61.6 & 51.7 & 76.3 & 6.9 \\
        Baselines                        &      &      &      &     \\
        \qquad -- Balanced binary tree   & 34.4 & 26.5 & 48.6 & 2.2 \\
        \qquad -- Random binary tree     & 28.5 & 22.0 & 40.3 & 1.1 \\
        \qquad -- Left-branching tree    & 30.8 & 23.8 & 43.6 & 0.5 \\
        \qquad -- Right-branching tree   & 36.0 & 27.8 & 50.9 & 0.4 \\
        \bottomrule
    \end{tabular}
    \caption{\textbf{Quantitative evaluation of the learned structure.} Similar to the evaluations used in constituency parsing \protect\cite{black1991procedure}, we report F$_1$ scores, precisions, recalls, and exact match (\%) rates with respect to a gold standard of Chinese character decomposition.}
    \label{tab:parsing_evaluation}
    \vspace{-2em}
\end{table}

\begin{figure*}[t]
    \centering
    \includegraphics[width=\linewidth]{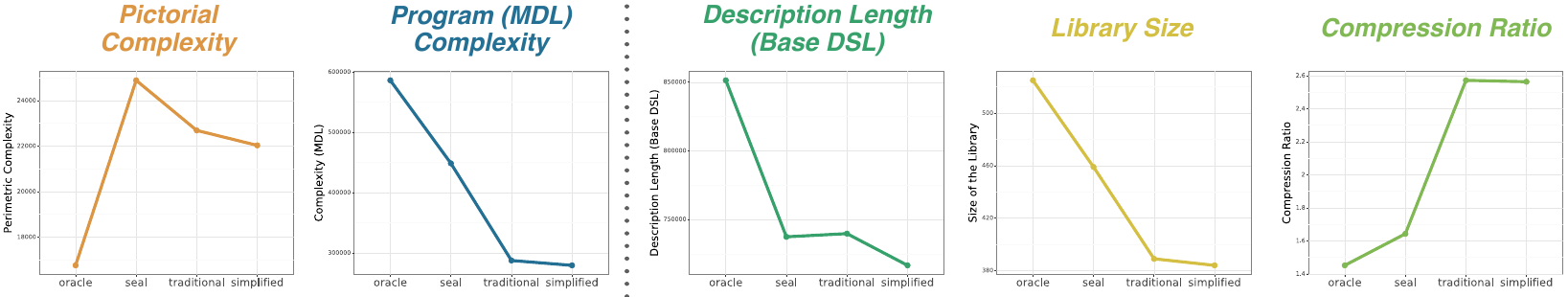}
    \vspace{-1.4em}
    \caption{\textbf{Changes in the quantifiable metrics over time.} We visualize \textbf{Left:} \textcolor{black}{pictorial complexity} (following \protect\citeA{han2022simplification}), \textcolor{black}{program (\ac{mdl}) complexity} calculated by our model \textcolor{mdlc}{$C(\mathcal{W})$}; \textbf{Right:} \textcolor{black}{description length under the base library} \textcolor{dlc}{$\mathrm{DL}_{\mathcal{L}_{base}}(\mathcal{W})$}, \textcolor{black}{learned library size} \textcolor{lsc}{$\lvert \mathcal{L}_{*} \rvert$}, and \textcolor{black}{compression ratio} \textcolor{crc}{$\mathrm{DL}_{\mathcal{L}_{base}}(\mathcal{W}) / \mathrm{DL}_{\mathcal{L}_{*}}(\mathcal{W})$} for the four scripts (oracle bone, seal, traditional, and simplified) respectively.}
    \label{fig:evolution_change}
    \vspace{-0.2em}
\end{figure*}

\myparagraph{Library functions resemble expert-defined radicals.} Prior research has shown that computational models of natural languages as hierarchical programs resemble linguists' theories in morphophonology \cite{goldsmith2001unsupervised,ellis2022synthesizing}. 
Can library learning models similarly uncover the structural theories underlying the Chinese language? 
In this section, we analyze the library functions learned by the model and compare them to expert-defined radicals in simplified Chinese.

Radicals are the graphical components that frequently occur in Chinese characters and have been used for indexing characters in dictionaries for almost two thousand years. Several seminal books have defined radical sets throughout history, notably, the Shuowen Jiezi Radicals (\zhtc{說文部首}, \cite{duan1821shuowen}), the Kangxi Radicals (\zhtc{康熙部首}, \cite{hanlin1885kangxi}), and the Table of Indexing Chinese Character Components (\zhsc{汉字部首表}, \cite{GF0011-2009}) developed by the Ministry of Education (MoE) in China. These radical sets were designed for different writing systems and varied in size. In the scope of this work, we leverage the MoE radicals set, as it is the only well-recognized radicals definition for the simplified Chinese writing system. 

The MoE radicals set contains 201 radicals: all radicals are parts of Chinese characters, and every character in Chinese has at least one identifiable radical from the MoE radicals set. Intuitively, these radicals can be interpreted as the most frequent co-occurring components discovered by experts in the simplified Chinese writing system and are designed to be easy to identify. Hence, our hypothesis is that our computational model would be able to rediscover most MoE radicals from the character programs.

By comparing the library functions learned by the model and the MoE radicals set, we found that our model discovered 187 ($93.0\%$) radicals in the MoE set. We illustrate the aligned MoE radical and library function pairs in \Cref{fig:moe_aligned}. 

\section{Part II: Complexity analysis across time}

Our results so far indicate that library learning effectively discovers structural patterns that link character representational efficiency and compression to their widely established prescriptive decompositions.
This alignment suggests that the model may accurately reflect human inductive biases toward combinatorial structure. 
If this premise holds, the model should reveal a gradual simplification as systems adapt to these biases in cultural evolution \cite{smith2003iterated}.

To test this prediction, the second part of our study presents a diachronic analysis of the Chinese writing system. To our knowledge, the only attempt to date to quantify script complexity of the Chinese at different historic stages comes from \citeA{han2022simplification}, which used a simple measure of pictorial complexity. Contrary to expectations, their findings did not support a trend towards simplification (\Cref{fig:evolution_change} left). 
However, the reliance on pictorial complexity, which fails to take into account systematic reuse across and within characters, could inadvertently inflate perceived character complexity.
By contrast, our model assesses writing system complexity on a holistic level rather than at the individual character level, capitalizing on the advantages of systematic compression through part reuse, and may thus provide a more accurate picture of writing system complexity.

\subsection{Methods and data}
\myparagraph{Data used for oracle bone, seal, traditional, and simplified.}
To test whether the Chinese writing system has indeed become simpler when taking combinatorial reuse into account, we analyze Chinese scripts at four representative historical stages \cite{kane2006chinese}: oracle bone ($\sim$1300 B.C.E), seal ($\sim$200 B.C.E), traditional ($\sim$5th century--present), and simplified (1956--present).

754 aligned characters were retrieved in each of these four stages of the Chinese writing system. For seal scripts, we used the correspondence data from the Unicode Seal Script Encoding Project \cite{wg2n5191}. For oracle bone scripts, the \zhtc{殷墟甲骨文字詞表} \cite<\textit{Yin Ruins Oracle Bone Script Lexicon},>[]{chen2010oracle,chen2012obi} was utilized for both the correspondence data and character shapes. 

Next, we obtained the program representations for all four scripts. For traditional and simplified Chinese, we adopted stroke decompositions from the Han character library.
Since there were no off-the-shelf stroke decompositions for seal and oracle bone characters. Thus, we manually labeled all strokes with a graphics tablet. Next, following \citeA{lake2011one} on defining stroke library, the recorded stroke trajectories were fitted into 2D Cubic Bézier curves and discretized the strokes into 33 primitives with K-means clustering, resulting in \acp{dsl} of the same size across the four scripts. Finally, we apply the same compression objectives and the iterative library learning algorithm as defined in Part I.

\vspace{-1em}
\myparagraph{Data used for traditional and simplified comparison.}
3,762 traditional-simplified Chinese character pairs were collected from \zhtc{標準字與簡化字對照手冊} \cite<\textit{Comparison Manual of Traditional and Simplified Chinese Characters,}>[]{moec2011} and the Han character library.

\subsection{Results}
\myparagraph{The Chinese writing system has been simplified over time.}
We applied the library learning model to the four scripts (oracle bone, seal, traditional, and simplified) of the Chinese writing system, respectively, and compared the program complexity metrics as defined in \cref{eq:objectives}. The results for the writing system's complexity $C(\mathcal{W})$, literal description length $\mathrm{DL}_{\mathcal{L}_{base}}$ (written in the base \ac{dsl} without library learning), learned library size $\lvert\mathcal{L}_*\rvert$, and compression rate $\mathrm{DL}_{\mathcal{L}_{base}}(\mathcal{W}) / \mathrm{DL}_{\mathcal{L}_{*}}(\mathcal{W})$ are shown in \Cref{fig:evolution_change}.

We observed a non-monotonic change in the literal description lengths with an overall trend of decreasing. The literal description length is practically a measurement of the total number of strokes used in the writing system. Oracle bone scripts were observed the largest literal description lengths, and there was an increase from the transition from seal script to traditional Chinese. 

To validate the simplification hypothesis we proposed earlier in the paper, we analyzed the complexity change over time. 
As shown in \Cref{fig:evolution_change} left, on the contrary to pictorial complexity measurement based on perimetric calculation used in drawing \citeA{han2022simplification}'s conclusion (a complexity ranking of seal $>$ traditional $>$ simplified $>$ oracle bone), we generally found that the writing system's program complexity $C(\mathcal{W})$ has shown a monotonic decrease across time (oracle bone $>$ seal $>$ traditional $>$ simplified), confirming earlier empirical arguments \cite{woon1987chinese,sampson1985writing,wang1973chinese,zhao2007planning}.

\vspace{-1em}
\myparagraph{Simplified Chinese is simpler but less systematic compared to traditional Chinese.}
The ``simplification'' from traditional to simplified Chinese was carried out as a deliberate process by the People's Republic of China between 1955 and 1986. 
Initially designed to ease the difficulty of learning and improve literacy by reducing strokes in individual characters \cite{chen1999modern,rohsenow2004fifty}, this process may have disrupted established systematicity and lead to a loss of established semantic-phonetic and graphic patterns \cite{handel2013logographic,zhao2011simplifying}. 

It has been suggested that this may have contributed to recognition and learning behavior differences \cite{liu2016transfer,mcbride2005chinese}.

Here, our model allows us to directly assess the systematicity of the traditional and simplified Chinese scripts. From the \ac{mdl} perspective, systematic-patterned data should be more compressive. Therefore, the compression ratio can be viewed as a proxy for the systematicity of a script.

\begin{figure}
    \centering
    \includegraphics[width=0.97\linewidth]{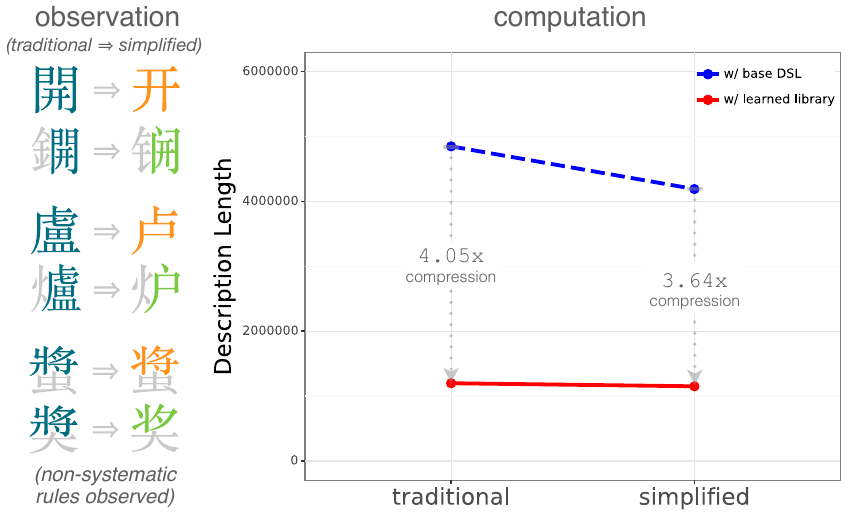}
    \caption{\textbf{Comparison of the compression ratio between traditional and simplified Chinese.} The traditional Chinese script is more compressible than simplified Chinese on the 3,762 aligned characters at a larger scale.
    }
    \label{fig:compression_tc_sc}
\end{figure}

By applying our model to the 3,762 aligned characters on the two scripts separately, we found (\Cref{fig:compression_tc_sc}) that although traditional has a noticeably larger description length under the base \ac{dsl} than simplified ($+16\%$), the gap was narrowed after compression ($+4\%$). Further analyzing the compression ratio, we observed that the traditional Chinese yielded a higher compression rate than the simplified Chinese, suggesting the simplification process from traditional to simplified Chinese did break part of the systematicity from a computational standpoint.

\section{Discussion}
In this work, we develop a library learning-based computational model, positing it as a framework for understanding the inductive biases behind the emergence and evolution of combinatorial structures in human language. 

The model is centered on the idea that combinatoriality develops from a \ac{mdl} perspective of representational efficiency, both by discovering inventories of reusable parts and by compressing the language using those parts.
This validity of this approach is demonstrated in the first part, where applying our model to the Chinese writing system uncovers known linguistic primitives and character decompositions, that align with intuitive human understandings of these languages.

Results from the second part reveal that these inductive biases, when applied over time, lead to the development of increasingly simple and efficient systems, as demonstrated in an analysis of the Chinese writing system across several stages of historic development.

There are multiple challenges to refining this computational framework. For instance, we are using canonical decompositions of modern Chinese scripts (\ie, traditional and simplified) but manually-parsed sequences for ancient scripts (\ie, oracle bone and seal), which could lead to less systematic representations of ancient scripts. Meanwhile, the stroke-based representation may not faithfully recover spatial layouts. These issues can be addressed by using unsupervised image parsing algorithms and including spatial templates \cite{lake2015human,hu2011image}.

In future work, we intend to further scale our model to analyze the systematic structure to not only forms but a broader range of form-meaning mappings. We hope our work can provide insights into how to build a computational model to reverse-engineer how compression, efficiency, and structure shape human languages \cite{kirby2015compression,gibson2019efficiency,tamariz2015culture,zaslavsky2018efficient}. More broadly, how large-scale library learning can contribute to theories of the evolution of efficient communicative systems and modeling human cognitive representations of abstractions.

\section{Acknowledgments}
We thank Yixin Zhu (Peking University), Robert Hawkins (UW-Madison), Judith Fan (Stanford), and Kevin Ellis (Cornell) for early discussions on this topic. Maddy Bowers, Alex Lew, Peng Qian (MIT), Yi Bai (Zi-tools) for many valuable discussions and contributions. We thank anonymous reviewers for their valuable comments.

MH, JM, LW, and JBT are supported by MIT Quest for Intelligence, AFOSR Grant \#FA9550-19-1-0269, the MIT-IBM Watson AI Lab, ONR Science of AI, and Simons Center for the Social Brain.

\newpage

\bibliographystyle{apacite}

\setlength{\bibleftmargin}{.125in}
\setlength{\bibindent}{-\bibleftmargin}

\bibliography{references}

\end{document}